\documentclass{article}

\usepackage{arxiv}

\usepackage[utf8]{inputenc} 
\usepackage[T1]{fontenc}    
\usepackage{hyperref}       
\usepackage{url}            
\usepackage{booktabs}       
\usepackage{amsfonts}       
\usepackage{nicefrac}       
\usepackage{microtype}      
\usepackage{lipsum}
\usepackage{svg}
\usepackage{graphicx}
\graphicspath{ {./images/} }
\usepackage{amsmath} 
\usepackage{pdflscape}
\usepackage{xcolor}
\usepackage[normalem]{ulem}
\usepackage{comment}
\usepackage{longtable}
\usepackage{tabularx}
\usepackage{caption} 
\usepackage{subcaption}
\usepackage{longtable}
\usepackage{booktabs}
\usepackage[utf8]{inputenc}
\usepackage{changepage}
\usepackage{amssymb} 

\usepackage{array}
\usepackage{float}
\newlength{\extralength}
\newlength{\fulllength}
\newcolumntype{C}{>{\centering\arraybackslash}X}

\title{Fall Detection for Smart Living using YOLOv5}

\author{Gracile Astlin Pereira \\[1ex]
\begin{minipage}[t]{0.90\textwidth}
\centering
\scriptsize Department of Computer Science, Huddersfield University, Queensgate, Huddersfield HD1 3DH, UK \\
Correspondence: U2292824@unimail.hud.ac.uk
\end{minipage}}

\begin{document}

\maketitle
\begin{abstract}
This work introduces a fall detection system using the YOLOv5mu model, which achieved a mean average precision (mAP) of 0.995, demonstrating exceptional accuracy in identifying fall events within smart home environments. Enhanced by advanced data augmentation techniques, the model demonstrates significant robustness and adaptability across various conditions. The integration of YOLOv5mu offers precise, real-time fall detection, which is crucial for improving safety and emergency response for residents. Future research will focus on refining the system by incorporating contextual data and exploring multi-sensor approaches to enhance its performance and practical applicability in diverse environments.
\end{abstract}

\keywords{Computer Vision; CNNs; Fall Detection; Object Detection; Real-Time Image processing; Smart Homes; YOLO; YOLOv5} 

\section{Introduction}

In the rapidly evolving domain of smart home technologies, prioritizing the safety and well-being of residents—particularly the elderly and individuals with mobility impairments—has become a significant concern. Falls are a critical risk factor, given their potential to cause severe injuries and long-term health issues. The integration of advanced technologies, such as artificial intelligence (AI) and the Internet of Things (IoT), has revolutionized the concept of smart homes, offering sophisticated solutions to enhance resident safety and quality of life. As such, effective fall detection systems are essential for improving emergency response times and mitigating the adverse effects of fall-related incidents \cite{zhu2022ethical, facchinetti2023can}.

Traditional fall detection solutions, such as wearable devices and home-installed sensors, often face limitations in terms of user acceptance and practicality. Wearable devices can be uncomfortable or prone to misuse, while fixed sensors may require significant infrastructural adjustments. These challenges underscore the need for more sophisticated, non-intrusive approaches to monitoring that seamlessly integrate into the home environment \cite{hu2023application, arar2021analysis}.
Among various vision-based techniques, the You Only Look Once (YOLO) algorithm has garnered significant attention for its real-time object detection capabilities and impressive accuracy. YOLO's ability to perform detection in a single pass of the network allows for rapid processing, which is crucial in dynamic environments like smart homes \cite{haritha2022real}. YOLOv3 \cite{zhao2020object} and YOLOv4 \cite{wu2020using}, early iterations of the algorithm, have made substantial contributions to object detection ~\cite{hussain2023review,aydin2023domain}. However, they have encountered challenges such as detecting small objects, managing occlusions, and maintaining processing speed, which can limit their effectiveness in complex and cluttered scenes.

The advancements in subsequent YOLO models have addressed many of these issues. YOLOv5 introduced significant improvements, including enhanced performance and efficiency detecting small targets, making it more suitable for real-time applications \cite{li2021yolo}. YOLOv6 \cite{li2022yolov6} further refined these capabilities with optimized network architectures that better handle diverse and challenging environments. YOLOv8 continued this trend, incorporating advanced features like multi-scale detection and improved occlusion handling. Despite these advancements, YOLOv8's increased complexity and resource demands may not always align with the constraints of certain deployment scenarios \cite{terven2023comprehensive}.

In this context, YOLOv5mu is an ideal choice for fall detection systems in smart homes due to its balance of accuracy and computational efficiency. As a lightweight variant of YOLOv5, YOLOv5mu offers effective real-time processing with a reduced model size, making it well-suited for edge devices and resource-constrained environments. This paper explores how YOLOv5mu can be utilized to develop an advanced fall detection system for smart homes, emphasizing its architecture, advantages, and practical implementation. The goal is to enhance fall detection technology and improve safety in modern residential settings.

\section{Literature Review} 

An innovative fall detection system using YOLO (You Only Look Once) object detection is introduced, designed to identify falls in real-time, particularly in low-light conditions. Achieving a notable accuracy rate of 93.16\%, the method leverages a pre-trained YOLO model and integrates image processing techniques for improved performance. Kamble et al. \cite{kamble2021fall} emphasize the importance of proper camera positioning, suggesting a 5:6 ratio between camera height and the distance of the person for effective detection. Despite relying on specific assumptions and camera setups, the system shows significant potential for real-time applications. Future enhancements are proposed, including adaptation for elderly care and accuracy improvements through advanced methods.

Wang and Jia \cite{wang2020human}, in their study, make a significant contribution to the field of fall detection by introducing an advanced method utilizing the YOLOv3 algorithm. Their approach innovates by optimizing YOLOv3 with K-means clustering for anchor boxes, which notably improves both accuracy and processing speed in detecting falls within complex environments. Their results, with an mean Average Precision (mAP) of 0.83, indicate a clear advancement over traditional algorithms. However, the study does not address potential limitations such as the algorithm’s performance across diverse environmental conditions or its adaptability to different populations. 

Long et al. \cite{long2021image} present a novel approach to fall detection for elderly individuals by integrating the YOLOv3 object detection algorithm with an image-based fall detection framework. The methodology involves capturing real-time video frames, applying YOLOv3 to detect and track individuals, and utilizing a fall detection algorithm to analyze posture changes and identify fall events. This system has been evaluated across various conditions, demonstrating up to 92\% accuracy in daylight and 60\% under low-light scenarios. Key advantages include its cost-effectiveness, seamless integration with existing camera systems, and high detection accuracy under optimal conditions. Nonetheless, the system faces limitations such as decreased performance in low-light environments, challenges with accuracy when multiple people are present, and potential issues with real-time processing and false positives.

Raza et al. \cite{raza2022human} in their work utilise Tiny-YOLOv4 for human fall detection, chosen for its high accuracy and low inference time, making it suitable for real-time applications. The methodology involved manual annotation of a dataset with 1691 fall images and 1731 normal images, split into 70\% training and 30\% testing. Training was conducted on a Titan XP GPU using various YOLO models, with Tiny-YOLOv4 achieving the best performance. Key metrics include a mAP of 95.2\%, precision of 95\%, recall of 96\%, F1-score of 95\%, and an inference time of 6.71 ms. The model was implemented on an edge device (OAK-D with Raspberry Pi 3) by converting YOLO weights to OpenVino format and using the DepthAI library for real-time inference, achieving an FPS of 26. This research stands out due to its real-time capability, edge implementation, high accuracy, and low inference time. It surpasses existing techniques using SVM, KNN, and other models, and suggests future enhancements with pose estimation and vision transformers, highlighting its potential for further advancements in fall detection systems.

In another study by Song, Yang, and Liu \cite{song2024optimizing} present significant advancements in fall detection technology. Unlike traditional models that rely on wearable sensors, environment-based methods, or earlier computer vision techniques which struggle with issues like occlusion and scene clutter, their study enhances YOLOv5 by integrating the CBAM and SE attention modules. These additions improve the network's ability to extract and represent critical features, while the Swish activation function replaces RELU to address gradient vanishing problems common in deep networks. As a result, the optimized model achieves an impressive 97.3\% accuracy on a dataset of 21,499 images, surpassing the performance of previous methods including those by Kwolek et al. \cite{kwolek2014human, kwolek2015improving}, Lahiri et al. \cite{lahiri2017abnormal}, and YOLOv5 with OpenPose, which showed lower accuracy and precision. Despite the challenge of requiring pre-labeled data, the approach marks a substantial improvement in real-time fall detection for public and healthcare environments.

Moutsis et al. \cite{moutsis2023fall} investigate the performance of YOLOv8 models for human detection, contrasting YOLOv8n with the more advanced YOLOv8s. Their study reveals that YOLOv8n, notable for its reduced computational requirements and inclusion of both original and rotated images during training, achieves superior accuracy and processing speed compared to YOLOv8s. Concurrently, Poonsri and Chiracharit \cite{poonsri2018improvement} assess various configurations of Multi-Layer Perceptrons (MLPs) and Support Vector Machines (SVMs), identifying YOLOv8n-r1 combined with SVM as providing an optimal balance of accuracy, sensitivity, and computational efficiency on a Raspberry Pi 4. Both Wang et al. \cite{wang2020fall} and Poonsri and Chiracharit \cite{poonsri2018improvement} underscore the necessity for integrating tracking mechanisms to enhance the precision of fall detection systems in dynamic real-world conditions.

Recent advancements in fall detection systems have greatly improved their accuracy and reliability, especially in elderly care. The introduction of YOLOv8, a state-of-the-art object detection framework, has led to a fall detection system with 90\% accuracy. This model surpasses previous versions like YOLOv3 and YOLOv5 by utilizing self-attention and multiscale technologies for real-time applications. It was fine-tuned with adjusted anchor box dimensions and a custom loss function to handle class imbalances, achieving a mAP of 90\% and an Area Under the Curve (AUC) of 0.894 on the precision-recall curve. The system consistently performs well across various Intersection-over-Union (IoU) levels, minimizing false positives \cite{krishnan2024robust}. Other applications of fall detection in industrial settings \cite{pereira2024fall} highlight its potential in enhancing safety and operational efficiency. Future advancements could focus on integrating posture detection and developing user-friendly interfaces, which would further enhance its real-world applicability, particularly in elderly care.

\section{Methodology}
\subsection{Dataset}
\begin{figure}[H]
\begin{adjustwidth}{-\extralength}{0cm}
\centering
\includegraphics[width=15cm]{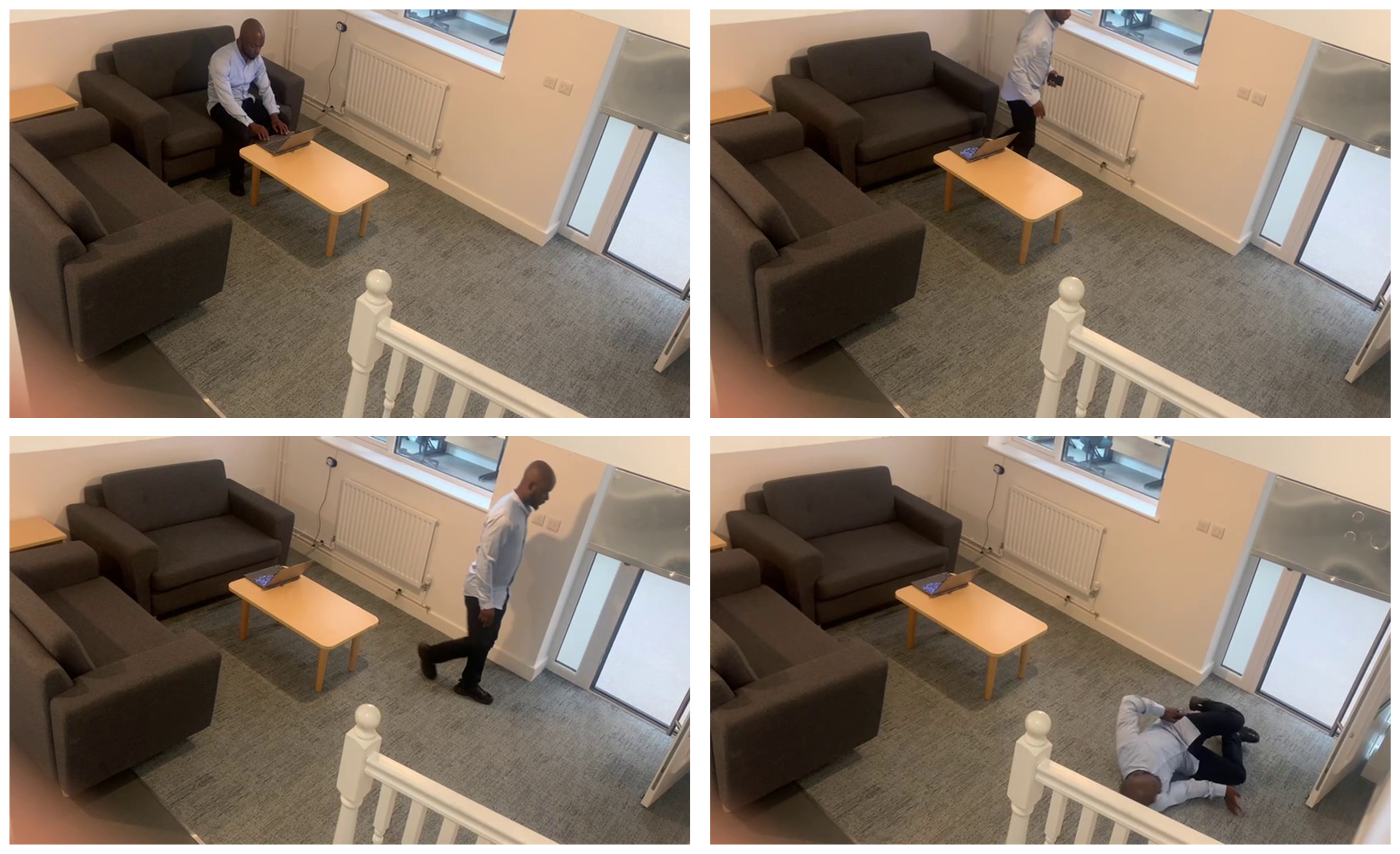}
\end{adjustwidth}
\caption{Smart Home Movement Dataset}
\label{Figure:1}
\end{figure}

The dataset was created from a high-resolution video recorded at 60 FPS with a resolution of 2560 x 1920 pixels using a phone camera. This video was processed with a Python script to extract 611 individual frames. These frames were manually annotated with Label Studio, and classified into four categories: Laptop, Occupant Sitting, Occupant Walking, and Occupant Abnormal (which indicates a Fall). A carefully chosen subset of 60 images from the full set was used for labeling. After manual annotation, the images underwent further augmentation using Roboflow. A selection of frames from the dataset is displayed in Figure \ref{Figure:1}.

\subsection{Data Augmentation}
Data augmentation plays a crucial role in enhancing model performance, particularly when working with a limited subset of images. Although the original dataset comprised 611 frames, only a selected subset of 60 images was used for annotation. To address the limited size of this subset and introduce variability, several augmentation techniques were applied using Roboflow. These included grayscale conversion to standardize color information, hue adjustment to vary color tones, saturation modifications to change color intensity, and brightness adjustments to alter lightness and darkness. By incorporating these augmentations, the dataset was diversified to better represent the range of colors and conditions present in the original video. This approach not only helps improve model performance across different color conditions but also reduces potential biases by ensuring the model is trained on a more representative and varied dataset.

\subsubsection{Random Resize}
\begin{figure}[H]
\begin{adjustwidth}{-\extralength}{0cm}
\centering
\includegraphics[width=15cm]{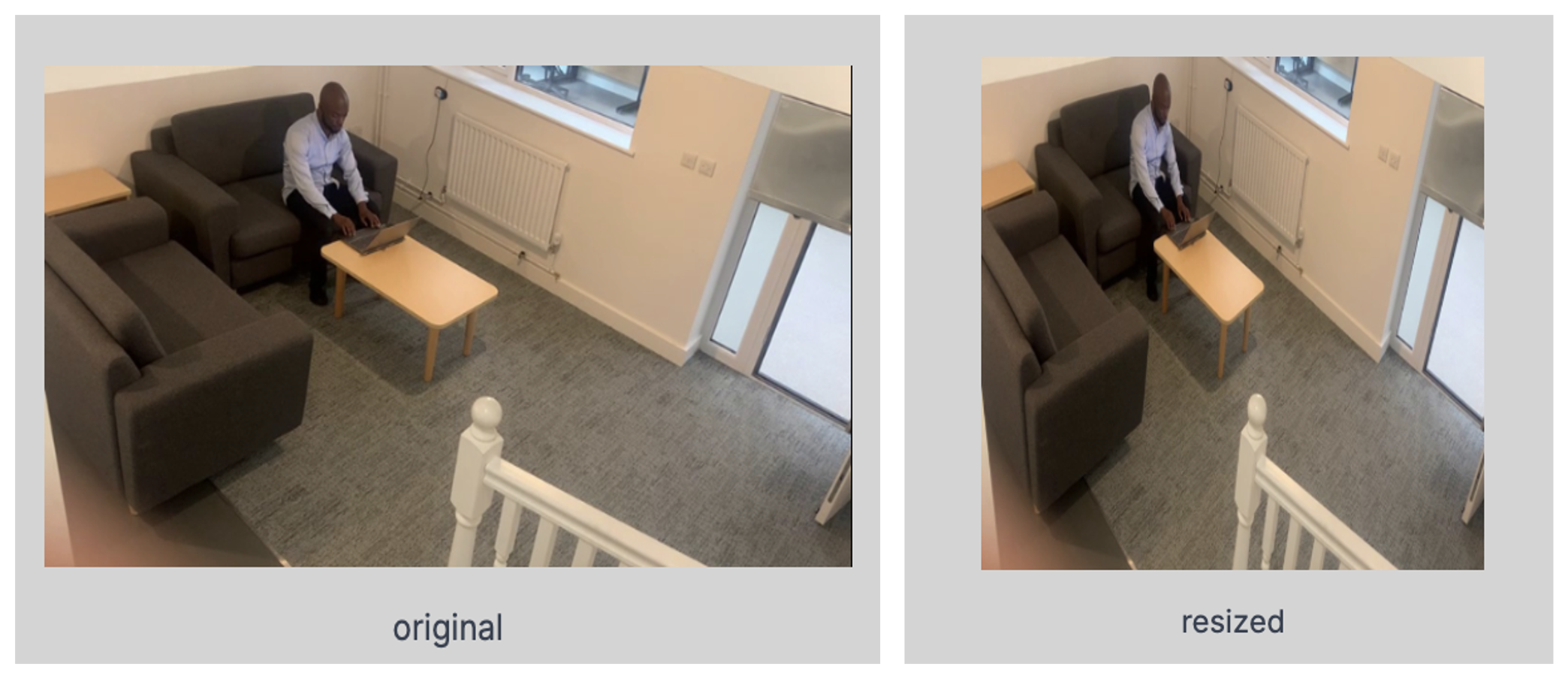}
\end{adjustwidth}
\caption{Image resized to 640x640 pixels}
\label{Figure:2}
\end{figure}

Resizing images is a crucial preprocessing step for training YOLO models, which are designed for real-time object detection. By standardizing the input dimensions from 2560 x 1920 pixels to 640x640 pixels, as illustrated in Figure \ref{Figure:2}, these models can effectively utilize this fixed input size to enhance their accuracy and efficiency in detecting and classifying objects within images. This resizing preserves spatial relationships and object features while providing a consistent input size that optimizes computational resources and enhances the model’s performance across various image scales.

\subsubsection{Random Grayscale}

The ToGray technique is applied to 15\% of the images, converting them to grayscale to simulate varied lighting conditions and reduce the model’s dependency on color information. This approach helps in creating a more diverse training dataset by introducing images with different intensity patterns. 

\begin{figure}[H]
\begin{adjustwidth}{-\extralength}{0cm}
\centering
\includegraphics[width=15cm]{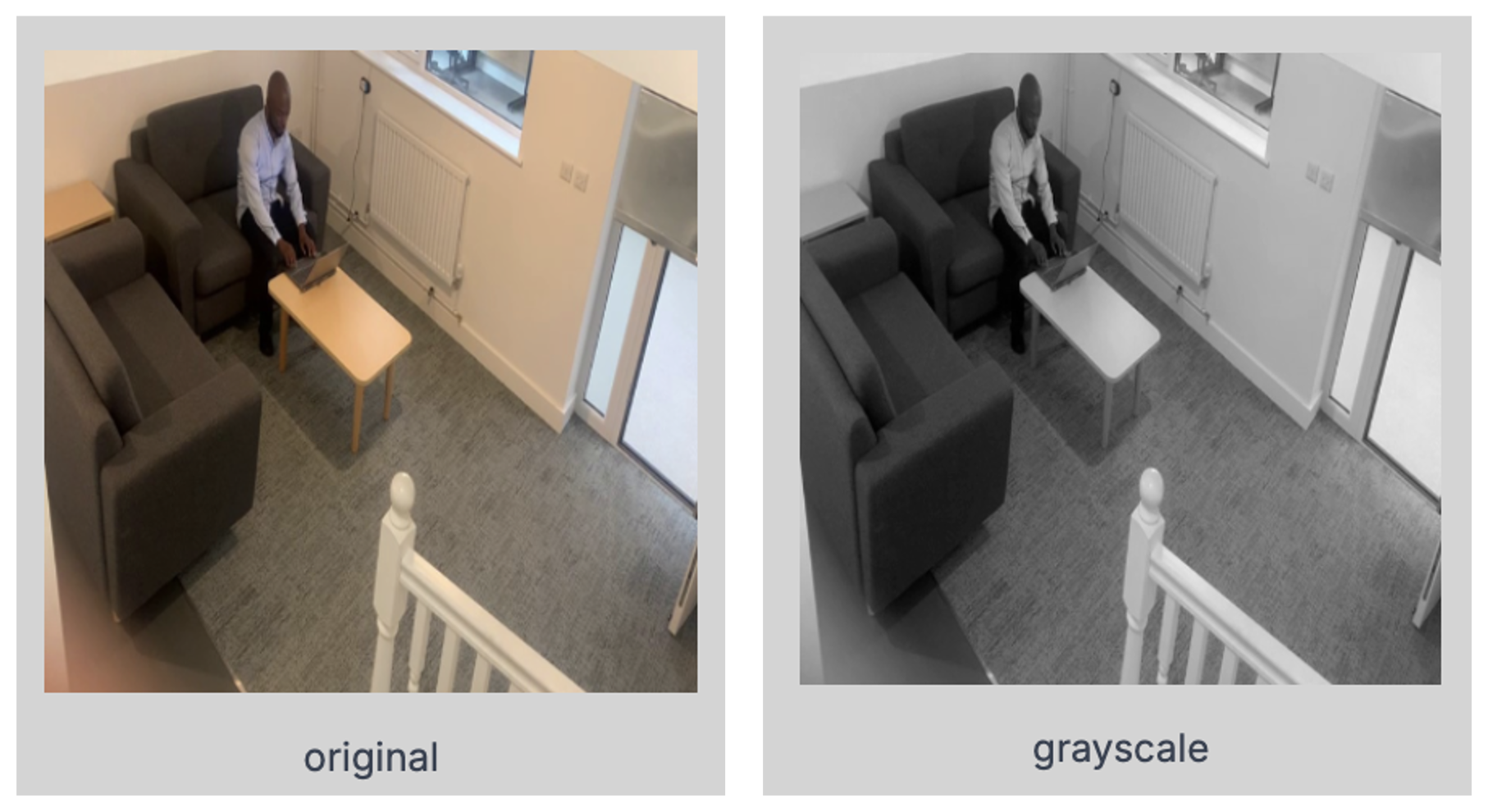}
\end{adjustwidth}
\caption{Original vs. Grayscale image}
\label{Figure:3}
\end{figure} 

An example of this augmentation is shown in Figure \ref{Figure:3}, which illustrates how grayscaling affects the images. By incorporating both color and grayscale images, the model learns to recognize features based on intensity, enhancing its robustness to various lighting scenarios. This method ultimately improves the model’s ability to detect hazards across diverse environments by making it less sensitive to color variations and more focused on structural patterns.

\subsubsection{Hue}
\begin{figure}[H]
\begin{adjustwidth}{-\extralength}{0cm}
\centering
\includegraphics[width=15cm]{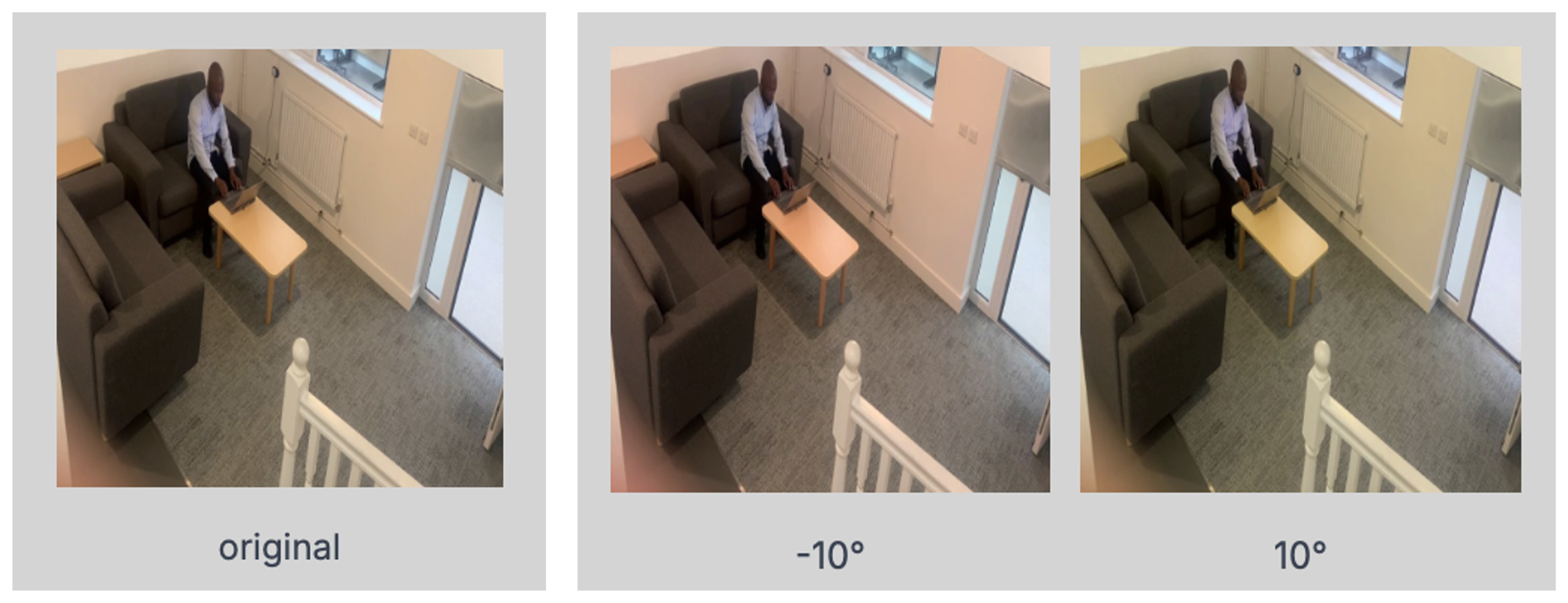}
\end{adjustwidth}
\caption{Images with hue adjustments}
\label{Figure:4}
\end{figure}

Applying hue adjustments of +10\% and -10\%, as demonstrated in Figure \ref{Figure:4}, involves systematically altering the chromatic properties of visual data by shifting the hue value on the color wheel. A +10\% hue adjustment results in a subtle shift of all color values toward a different hue, potentially enhancing or altering the perceived warmth or coolness of the colors. Conversely, a -10\% adjustment shifts the hues in the opposite direction, modifying the color spectrum to create a contrasting chromatic effect. These adjustments are instrumental in simulating varied lighting conditions and color environments within the dataset, thereby enhancing the robustness and adaptability of machine learning models trained on this data.

\subsubsection{Saturation}
\begin{figure}[H]
\begin{adjustwidth}{-\extralength}{0cm}
\centering
\includegraphics[width=15cm]{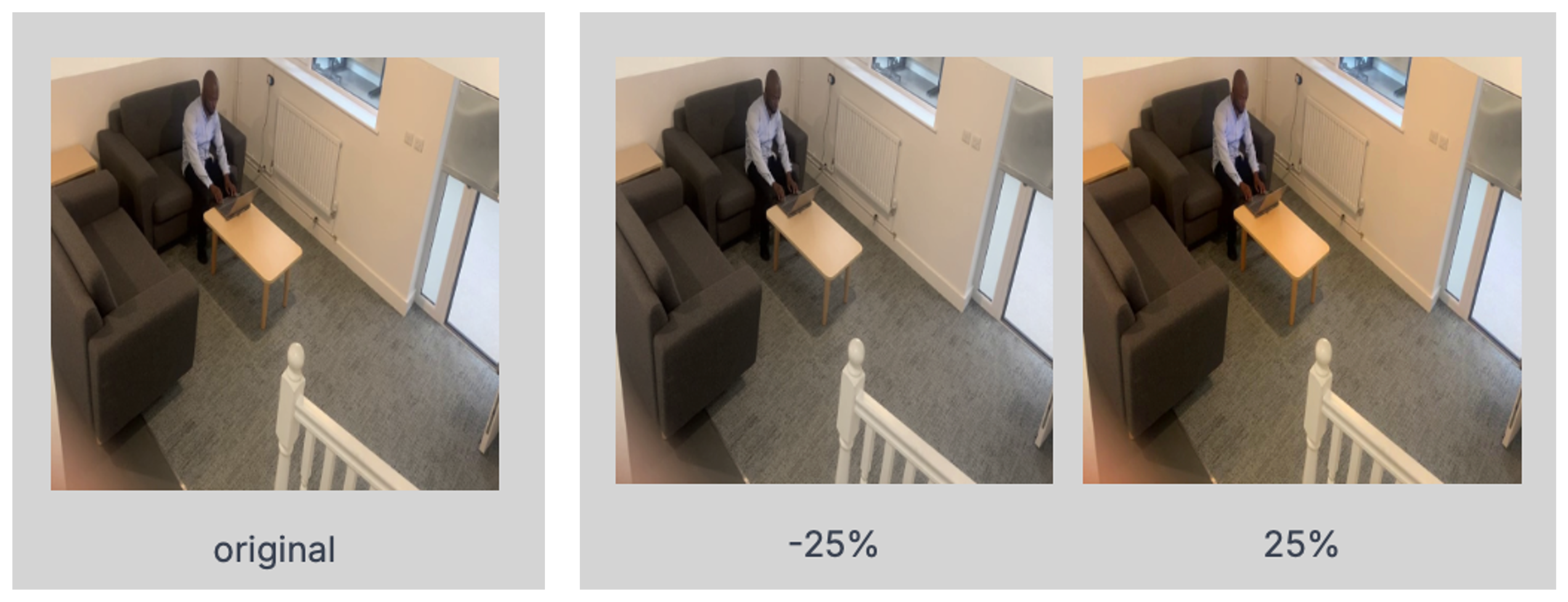}
\end{adjustwidth}
\caption{Saturation applied on dataset}
\label{Figure:5}
\end{figure}

Saturation adjustments of +25\% and -25\%, as illustrated in Figure \ref{Figure:5}, involve modifying the intensity of colors within visual data by altering their saturation levels. A +25\% saturation adjustment amplifies the chromatic intensity, rendering colors more vivid and pronounced. In contrast, a -25\% adjustment diminishes the saturation, resulting in more muted and desaturated color tones. These adjustments are crucial for simulating diverse environmental conditions, such as variations in lighting or color richness, thereby contributing to the robustness and generalization capabilities of machine learning models trained on this dataset.

\subsubsection{Brightness}
\begin{figure}[H]
\begin{adjustwidth}{-\extralength}{0cm}
\centering
\includegraphics[width=15cm]{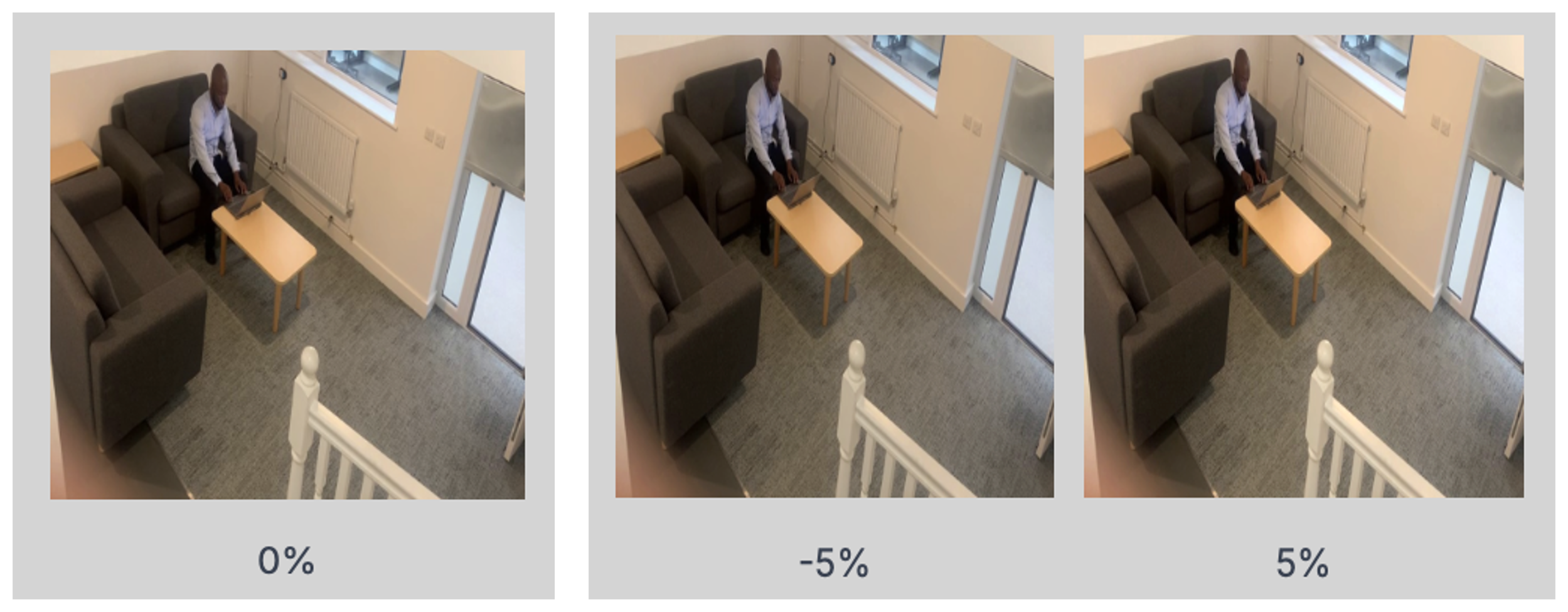}
\end{adjustwidth}
\caption{Images with brightness adjustments}
\label{Figure:6}
\end{figure}

Brightness adjustments of +5\% and -5\%, as demonstrated in Figure \ref{Figure:6}, involve altering the luminance of visual data by increasing or decreasing the overall lightness of the image. A +5\% brightness adjustment slightly enhances the light levels, making the image appear brighter and more illuminated. Conversely, a -5\% adjustment reduces the light levels, resulting in a darker, more subdued image. These brightness modifications are essential for simulating different lighting conditions within the dataset, thereby improving the robustness and adaptability of machine learning models to varying environmental factors.

\subsection{YOLOv5m Architecture}

The YOLOv5 model architecture is structured into three main components: the backbone, neck, and head. The backbone of YOLOv5 incorporates a ResNet-based cross-stage partial (CSP) network, which enhances efficiency by using cross-stage partial connections and includes multiple spatial pyramid pooling (SPP) blocks to capture various features while reducing computational demands. The neck of the model integrates a path aggregation network (PAN) module along with upscaling layers to improve the resolution of the feature maps and facilitate better information exchange across layers. Finally, the head of YOLOv5 consists of three convolutional layers that are responsible for predicting bounding boxes, class labels, and confidence scores \cite{xu2021aforest, hussain2024depth}.

The YOLOv5mu architecture initiates with a foundational convolutional layer, Conv1, which projects the input tensor from 3 channels to 48 channels, utilizing a kernel of size 6 × 2 with a stride of 2. This is succeeded by Conv2, which enhances the feature representation to 96 channels through a kernel size of 3 × 2, also with a stride of 2, facilitating further spatial down-sampling and feature extraction.

Following these initial stages, the model integrates a series of CSP blocks, exemplified by C3-1, which maintains 96 channels without modifying the spatial dimensions. This is followed by progressively deeper convolutional layers, such as Conv3 and Conv4, which expand the feature channels to 192 and 384, respectively. These layers leverage kernel sizes of 3 × 2 and strides of 2, contributing to the refinement of feature hierarchies while diminishing spatial resolution. Following C3 blocks further refine these features, while Conv5 scales up the output channels to 768. C3-4 continues this process of feature extraction without altering the spatial resolution.

A pivotal component, the Spatial Pyramid Pooling (SPPF) layer, performs multi-scale feature aggregation while preserving a fixed spatial dimension of 20 × 20 pixels. This is followed by a convolutional layer (Conv6) that reduces the feature channels to 384. The architecture then employs upsampling operations to enhance the spatial resolution, with the Concat layer fusing features from various scales and stages, thus enriching the feature representations.

Subsequent layers, including Conv7 and Conv8, adjust channel dimensions and spatial resolution, while further concatenation and processing through C3-5 and C3-5 blocks refining the feature maps. The network culminates with Conv9 and C3-7, which enhance high-dimensional features.

\begin{table}[H]
\renewcommand{\arraystretch}{1.3}
\setlength{\tabcolsep}{15pt} 
\caption{YOLOv5mu Architecture}
\label{tab:yolov5mu}
\centering
\begin{tabular}{lcccc}
\hline
\textbf{Layer} & \textbf{Input Channels} & \textbf{Output Channels} & \textbf{Size} & \textbf{Parameters} \\
\hline
Conv1 & 3 & 48 & 6 × 2, stride 2 & 5280 \\
Conv2 & 48 & 96 & 3 × 2, stride 2 & 41664 \\
C3-1 & 96 & 96 & - & 65280 \\
Conv3 & 96 & 192 & 3 × 2, stride 2 & 166272 \\
C3-2 & 192 & 192 & - & 444672 \\
Conv4 & 192 & 384 & 3 × 2, stride 2 & 664320 \\
C3-3 & 384 & 384 & - & 2512896 \\
Conv5 & 384 & 768 & 3 × 2, stride 2 & 2655744 \\
C3-4 & 768 & 768 & - & 4134912 \\
SPPF & 768 & 768 & 5 × 5 pooling & 1476864 \\
Conv6 & 768 & 384 & 1 × 1 & 295680 \\
Upsample & 384 & 384 & Upsample (2 × 2) & 0 \\
Concat & 384, 384 & 384 & Concat & 0 \\
C3-5 & 384 & 384 & - & 1182720 \\
Conv7 & 384 & 192 & 1 × 1 & 74112 \\
Upsample & 192 & 192 & Upsample (2 × 2) & 0 \\
Concat & 192, 192 & 192 & Concat & 0 \\
C3-6 & 192 & 192 & - & 296448 \\
Conv8 & 192 & 192 & 3 × 2, stride 2 & 332160 \\
Concat & 192, 192 & 192 & Concat & 0 \\
C3-7 & 192 & 384 & - & 1035264 \\
Conv9 & 384 & 384 & 3 × 2, stride 2 & 1327872 \\
Concat & 384, 384 & 384 & Concat & 0 \\
C3-8 & 384 & 768 & - & 4134912 \\
Detect & 192, 384, 768 & - & Detect (anchors) & 4220380 \\
\hline
\end{tabular}
\end{table}

YOLOv5, anchor-based predictions link bounding boxes to predefined anchor boxes of specified dimensions. The loss function integrates Binary Cross-Entropy for class and objectness losses, and Complete Intersection over Union (CIoU) for location loss, with objectness loss varying by prediction layer size \cite{liu2023lightweight, hussain2024yolov1}. YOLOv5mu's detailed architecture, including filter counts, sizes, and layer repetitions, is outlined in Table \ref{tab:yolov5mu}. 

The final stages of the network include C3-8, which further processes the features, culminating in the Detect layer. This layer, crucial for anchor-based object detection, integrates multi-scale feature maps to produce bounding box predictions, class labels, and confidence scores, optimizing detection performance through sophisticated feature aggregation and prediction strategies.

\section{Results}

\begin{figure}[H]
\begin{adjustwidth}{-\extralength}{0cm}
\centering
\includegraphics[width=15cm]{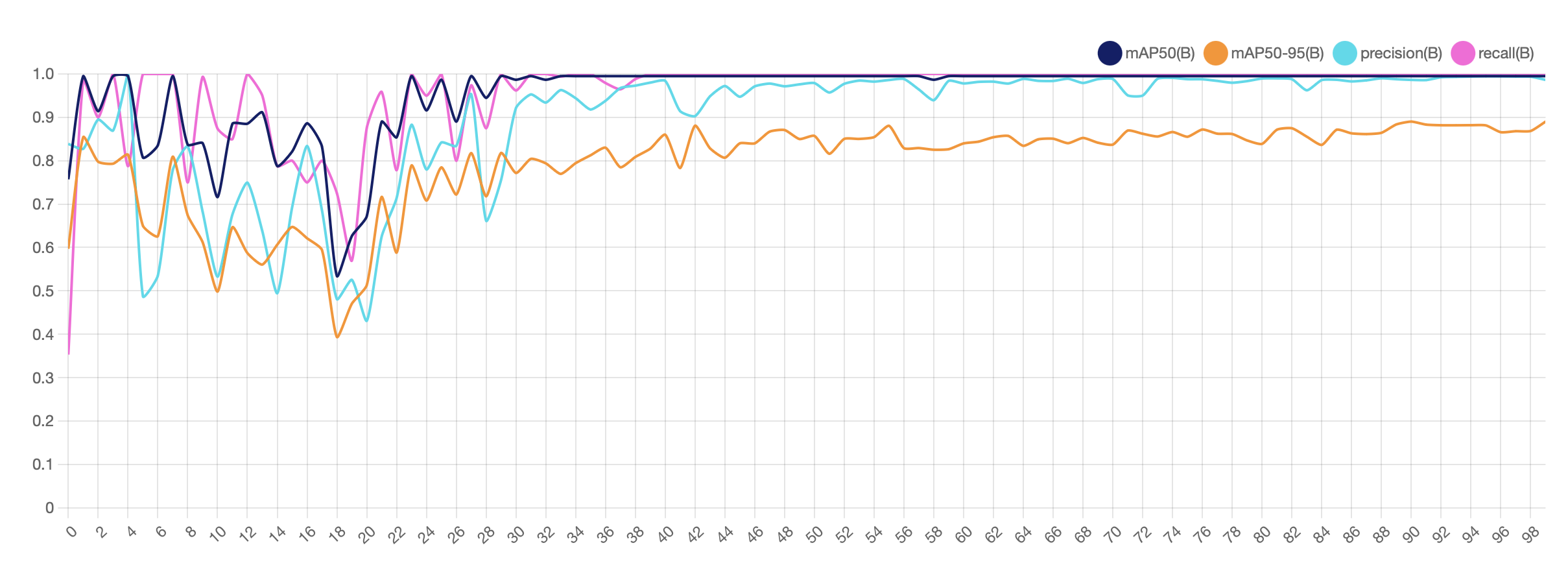}
\end{adjustwidth}
\caption{Performance for YOLOv5mu Model}
\label{Figure:7}
\end{figure}

The YOLOv5mu model, trained for 100 epochs on a Tesla T4 GPU, demonstrates impressive performance in object detection for the smart home dataset, as detailed in Figure \ref{Figure:7}. Throughout the training, the model exhibited consistently strong results, with notable performance metrics across all classes.

In the final validation phase, the model achieved an overall precision of 0.986 and a recall of 1.000, reflecting its effectiveness in accurately identifying objects and minimizing omissions. The mean average precision at a 50\% IoU threshold (mAP50) reached 0.995, indicating reliable and precise detections. Meanwhile, the mean average precision averaged over multiple IoU thresholds (mAP50-95) was 0.890, showcasing robust performance even with varying object overlaps.

For specific classes, the model's performance was exceptional. In the "Laptop" category, precision and recall were both perfect at 1.000, with an mAP50 of 0.995 and an mAP50-95 of 0.943, reflecting near-ideal detection capabilities. The "Occupant State - Abnormal" class demonstrated a precision of 0.971 and a recall of 1.000, with high mAP50 (0.995) and mAP50-95 (0.895) scores, indicating effective detection. Similarly, the "Occupant State - Sitting" class achieved a precision of 0.986 and a recall of 1.000, supported by high mAP50 (0.995) and mAP50-95 (0.946) scores.

For the "Occupant State - Walking" class, the model exhibited high precision at 0.988 and perfect recall, but with a slightly lower mAP50-95 of 0.776. This suggests that while the model effectively detects walking states, there may be some challenges in maintaining this high performance across different IoU thresholds.

\begin{figure}[H]
\begin{adjustwidth}{-\extralength}{0cm}
\centering
\includegraphics[height=11.5cm]{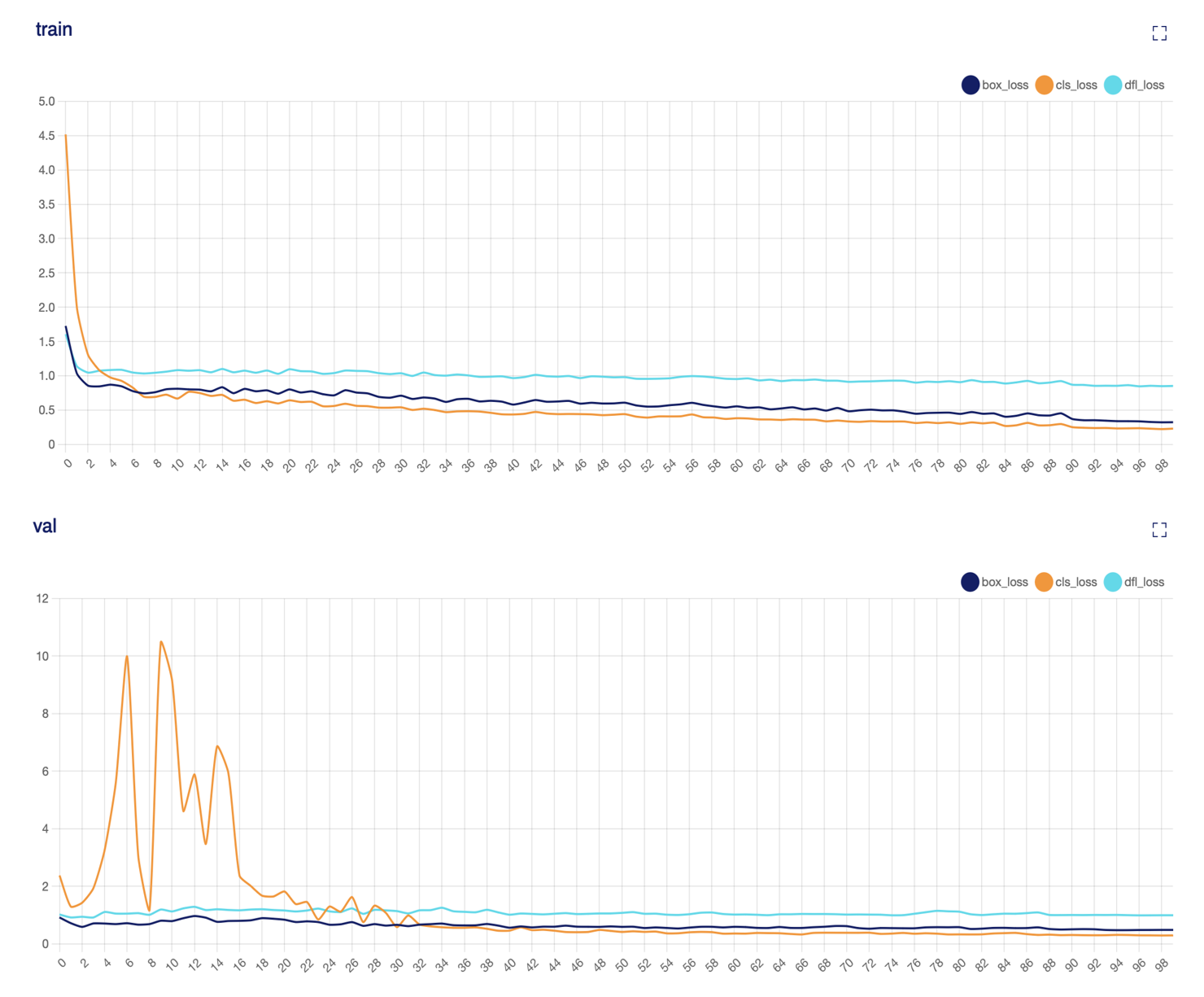}
\end{adjustwidth}
\caption{Training and Validation Loss Curves - YOLOv5mu}
\label{Figure:8}
\end{figure}

The loss analysis over the 100 epochs of training, depicted in Figure \ref{Figure:8}, reveals the model's progress and refinement in learning object detection. At the final epoch, the box loss is recorded at 0.323. This figure reflects the model's ability to accurately predict the locations of objects. Although there is a slight increase from earlier epochs, the value remains low, suggesting that the bounding box predictions are relatively stable and precise by the end of training.

The classification loss stands at 0.2295 at epoch 100. Classification loss measures how well the model identifies and differentiates between various classes. This relatively low classification loss indicates that the model has become proficient in distinguishing between the classes with minimal errors, thereby improving its classification accuracy.

The distribution focal loss (DFL), which assesses the model's accuracy in predicting object boundaries, is recorded at 0.8519. The DFL shows a modest increase from previous epochs, which is generally expected as the model fine-tunes its boundary predictions. Despite this increase, the value remains controlled, implying that the model's boundary predictions are becoming increasingly accurate and that the learning process has been successful in refining this aspect of detection.

Overall, the loss metrics demonstrate that the model has achieved a well-balanced state of learning, with minimal loss values indicating effective training. The stability in the loss figures across epochs suggests that the model's training has been successful in minimizing errors in object localization and classification, leading to robust and reliable performance. This architecture has utilized 339 layers and a total of 25,067,452 parameters. The model operates with 64.0 GFLOPs, reflecting a balance between computational efficiency and detection capability. This setup ensures that the model can perform high-precision detection tasks while maintaining reasonable computational demands.

\section{Conclusion}

The YOLOv5mu model, trained on the smart living dataset, has proven to be a highly effective solution for fall detection, achieving an impressive mean average precision (mAP) of 0.995. This exceptional performance demonstrates the model's ability to accurately identify fall events among various objects and activities, which is crucial for enhancing resident safety and emergency response. In addition to its primary application in fall detection, the model's strong performance in object detection underscores its versatility in identifying potential hazards within a smart home environment. The use of data augmentation techniques has further enhanced the model's robustness and adaptability.

Future work could explore integrating additional contextual information, such as posture analysis and activity recognition, to further improve fall detection accuracy and differentiate between various types of falls and other hazards. Enhancing the model's capabilities to operate in diverse lighting conditions and adapting it to different home environments could also be valuable ~\cite{hussain2023child}. Moreover, investigating the application of other YOLO models in conjunction with other sensor modalities, such as wearable devices or environmental sensors, may provide a more comprehensive safety solution, expanding its utility and effectiveness in real-world smart home settings as well as other industrial domains ~\cite{alif2024lightweight}.

\begin{adjustwidth}{-\extralength}{0cm}

\bibliographystyle{unsrt}  
\bibliography{references}  

\end{adjustwidth}
\end{document}